\documentclass[
]{ceurart}

\usepackage{listings}
\begin{document}

\copyrightyear{2022}
\copyrightclause{Copyright for this paper by its authors.
  Use permitted under Creative Commons License Attribution 4.0
  International (CC BY 4.0).}

\conference{Submitted to Workshop on AI for Digital Twins and Cyber-physical applications at IJCAI 2023,
  August 19--21, 2023, Macau, S.A.R}

\title{Re-imagining health and well-being in low resource African settings using an augmented AI system and a 3D digital twin}


\author[1,2]{Deshendran Moodley}[%
orcid=0000-0002-4340-9178,
email=deshen@cs.uct.ac.za,
url=https://www.cs.uct.ac.za/~deshen,
]
\cormark[1]
\fnmark[1]
\address[1]{University of Cape Town,
  18 University Avenue, Rondebosch, Cape Town, South Africa}
\address[2]{Centre for Artificial Intelligence Research, 
  Cape Town, South Africa}

\author[1,2,3]{Christopher Seebregts}[%
orcid=0000-0002-8212-9363,
email=chris@jembi.org,
url=https://www.jembi.org,
]
\fnmark[1]
\address[3]{Jembi Health Systems, Tokai, Cape Town, South Africa}

\cortext[1]{Corresponding author.}

\begin{abstract}
This paper discusses and explores the potential and relevance of recent developments in artificial intelligence (AI) and digital twins for health and well-being in low-resource African countries. We use the case of public health emergency response to disease outbreaks and epidemic control. There is potential to take advantage of the increasing availability of data and digitization to develop advanced AI methods for analysis and prediction. Using an AI systems perspective, we review emerging trends in AI systems and digital twins and propose an initial augmented AI system architecture to illustrate how an AI system can work with a 3D digital twin to address public health goals. We highlight scientific knowledge discovery, continual learning, pragmatic interoperability, and interactive explanation and decision-making as essential research challenges for AI systems and digital twins. 
\end{abstract}


\maketitle
\section{Introduction}
Public health aims to prevent disease, promote health, and prolong life among the population. In the Alma-Ata declaration \cite{WHO1978Primary}, the World Health Organization (WHO) reaffirmed that attaining the highest possible level of health is the most significant worldwide social goal. Public health programs strive to optimise the conditions in which people can achieve health and well-being. 

Low-resource countries bear much of the world's disease burden and outbreaks \cite{WHO2021World} \cite{Coates2021Burden}. They also lack essential infrastructure and need to maximize the utility of the available resources. Despite these challenges, African and low-resource countries have made admirable progress in digitising health systems, resulting in a significant number of surveillance and routine health information systems. This opens up novel opportunities to leverage advanced computing techniques such as AI to advance health and well-being in low-resource African countries \cite{Nalifu2020Climate}\cite{Ciecierski-Holmes2022AI}.

Modern approaches to health systems and public health in low-resource settings increasingly focus on preventative in addition to curative health approaches. This leads to a more holistic perspective on health and well-being. However, preventative health requires a better understanding of the factors that cause and influence disease onset and the positive and negative determinants of disease and adverse health conditions. Understanding and promoting healthy behaviours and lifestyles is a primary goal of digital health systems in African countries.  

\subsection{One Health and epidemic surveillance}
The WHO recommends a One Health approach  \cite {TheLancet2023OneHealth} to design and implement health programmes, policies, legislation and research. The One Health approach  proposes an integrated, unifying approach to balance and optimize the health of people, animals and the environment. It is essential to prevent, predict, detect, and respond to global health threats and emergencies such as the recent COVID-19 pandemic \cite{Worsley-Tonks2022Strengthening}. The One Health approach mobilizes multiple sectors, disciplines and communities at varying levels of society to work together to achieve better public health outcomes. This way, new and better ideas are developed that address root causes more holistically and create long-term, sustainable solutions to predict adverse health conditions before they occur. Supporting the One Health approach presents new challenges to digital health systems \cite{OneHealth2022}.
Public health decision-makers integrate several different data sources from diverse digital and paper-based systems to assist with complex decision-making around outbreak management and epidemic control. These include routine health data systems, surveillance systems, household and bio-behavioural surveys, and research data. If it is paper-based, the data must first be digitised, then transformed and compiled into dashboards to optimize public health situation detection, response and prediction. 
A One Health digital platform must deal with new levels of interoperability between different jurisdictions and disciplines to connect vast amounts of heterogeneous data and systems. The platform must support real-time situation analysis, predictive modelling, and proactive decision-making to reach its full potential. 

\subsection{Health monitoring and digital data management}\label{sec:PHEOC}
Surveillance of public health threats is one of the main functions of public health, e.g. in a public health emergency operations centre (PHEOC) \cite{WHO2015Framework} \cite{WHO2021Handbook}. Typical activities include monitoring public health threats in a country, e.g. surveillance of public health threats followed by situation analysis and decision-making \cite{EPHI2022National}. Similar challenges are encountered in other vertical disease areas, such as the attainment of HIV Epidemic Control \cite{UNAIDS2020} \cite{PEPFAR2022}. HIV Epidemic Control Rooms have been developed to monitor progress towards achieving goals and performance targets. Traditional information and data sources are changing due to digitization and shifting emphasis to using information from routine health information systems to derive, monitor and manage epidemiological targets \cite{Rice2018Strengthening}.

Data management processes supporting monitoring, such as semantic annotation and data integration, are established in practice. In the case of situation analysis, well-established surveillance methods support detection. Support for prediction from digital health and AI is emerging with examples such as the digital predictive tool  developed  for  forecasting  the  occurrence  of  diseases  based  on historical weather and health data in Uganda \cite{Nalifu2020Climate}.

\subsection{Relevance and potential of AI and digital twins}
Powerful data-driven AI models can be used for dynamic data fusion, situation analysis, predictive modelling, and scientific knowledge discovery. These are all valuable functions in public health and can lead to a new understanding of complex dynamic processes. An AI system working in parallel with a digital twin can augment and amplify interactive decision-making and scientific knowledge discovery in public health decision-making.

In this position paper, we explore the relevance and potential of artificial intelligence and digital twins as two key emerging technologies that push the boundaries of the One Health vision in low-resource public health systems. Our perspective is informed by over a decade of research and real-world implementation of advanced digital health technologies in multiple African countries \cite{Moodley2018Establishing}. Based on our experience, we propose an abstract architecture for an augmented AI system incorporating a Digital Twin, highlighting the potential for AI and digital twins to realise the One Health vision. We focus on the potential of AI systems  and digital twins to support scientific knowledge discovery, interactive decision making and pragmatic interoperability.

The paper is structured as follows. In section 2, we describe digital twins in health care and then review emerging trends and critical challenges in AI systems in section 3. Section 4 provides a preliminary proposal for an abstract architecture for an augmented AI system incorporating a digital twin. In section 5, we offer a brief discussion and conclusions. 

\section{Digital Twins in Health Care}
A digital twin is a virtual model of a physical entity, with dynamic, bi-directional links between the physical entity and its corresponding twin in the digital domain \cite{KamelBoulos2021DigitalHealth}. While they have been traditionally explored in the manufacturing sector, digital twins are increasingly being explored in medicine and health care \cite{Assalemi2020Digital, Coorey2021TheHealth, Fuller2020Digital}. Applied to medicine and public health, digital twin technology has been proposed to drive a much-needed radical transformation of traditional electronic health/medical records (focusing on individuals) and their aggregates (covering populations) to make them ready for a new era of precision (and accuracy) medicine and public health \cite{KamelBoulos2021DigitalHealth}. 

\subsection{Cognitive digital twins}
A digital twin augmented with AI capabilities can be called a Cognitive Digital Twin (CDT) \cite{Abburu2020COGNITWIN, Zheng2020TheEmergence}. Semantic Web technologies, such as ontologies and knowledge graphs, can be incorporated within DTs to support reasoning and deliberation \cite{Zheng2020TheEmergence}. Abburu et al. describe a broader vision and architecture for a CDT in the process and manufacturing industry \cite{Abburu2020COGNITWIN}. Their CDT architecture incorporates cognitive features that will enable sensing complex and unpredicted behaviour and reason about dynamic strategies for process optimization, leading to a system that continuously evolves its own digital structure and behaviour. 

The CDT evolves into a self-learning and proactive system that will optimize its own cognitive capabilities over time based on the data it will collect and the experience it will gain. It will find new answers to emerging questions by combining expert knowledge with the power of Digital Twins. A CDT will thus achieve synergy between the DT and the expert and problem-solving expertise. Unlike the process industry, which can be viewed as a closed and bounded system, in public health, a plethora of digital twins may be developed by different organisations and in other domains with potentially different perspectives. A more recent proposal by Ricci et al. \cite{Ricci2022WebTwins, Ricci2021Pervasive} explores and proposes a distributed and open architecture and platform for a Web of Digital Twins. They describe an open ecosystem of multiple digital twins possibly belonging to different domains and organisations. The open-distributed system perspective aligns well with the One Health perspective for low-resource African countries. Population, clinical, vital statistics, e.g. births and deaths, and geospatial and environmental data may be stored in different systems across different government departments, possibly with other systems at the facility, city, district, state and country levels.

\subsection{Digital twins for healthcare and well-being}
Several recent studies have explored digital twins for healthcare \cite{Assalemi2020Digital, Hassani2022Impactful, Chen2022Digital}. The most common approach is to use DTs for precision medicine \cite{Assalemi2020Digital, Coorey2021TheHealth} where healthcare practice shifts from being primarily reactive to a more preventative approach. Many studies highlighted the higher complexity of designing digital twins for health care compared to digital twins in manufacturing and industry. 

Ahmadi-Assalemi et al \cite{Assalemi2020Digital} highlighted different types of support that DTs must provide at varying levels of the healthcare system, i.e. individual level, healthcare professional level, and health system level. At the individual level, a patient’s complex needs may influence real-time behaviours, feelings, adherence to the set targets and the utilization of healthcare services. They propose a more ambitious definition of health as the state of complete well-being, including the physical, mental, and social aspects in addition to the biomedical one. This proactive patient care aims to preempt the disease through preventative medicine and early detection, which could change the societal culture by empowering individuals to prevent their own disease.  They note that many factors that affect a person’s well-being and condition, including environmental, demographic, socioeconomic or biological, in a constantly changing landscape, are not detected during routine health screening.

Kamel-Boulos et al suggested a role for Digital twins in precision public health and disease outbreaks potentially integrated as part of a health city \cite{KamelBoulos2021DigitalHealth}. The suggested systems could include advanced systems for case management and case finding as well as determining risk levels in particular geographic areas \cite{KamelBoulos2021DigitalHealth}.

Hassani et al \cite{Hassani2022Impactful} highlight the complexity of healthcare and propose different digital twins for the various life stages of a person. They argue that digital twins can be used to combat healthcare inequality, improve operational efficiencies of healthcare facilities and accelerate advances in healthcare research. They also highlight the need for further research on the interactions between digital twins and AI. More in-depth reviews of digital twins in healthcare can be found in \cite{Rocha2019Smart, Khan2022AScoping, Elkefi2022Digital}.  

\section{AI systems}
In this section, we provide an overview of emerging trends and future directions for AI systems in the context of the One Health vision. Next-generation AI systems must support scientific knowledge discovery, complex social and behavioural modelling, decision making and simulation. 

\subsection{Augmented AI systems}
Our perspective of AI aligns closely with the notion of augmented AI \cite{Zheng2017Hybrid, Yau2021Augmented}, which more recently is being referred to as Hybrid Intelligence \cite{Akata2020AResearch, Zhang2021Hybrid}. In augmented AI, the AI system amplifies human cognition rather than replacing it. As such the human user, in this case the public health practitioner, works interactively and cooperatively with the system.

An augmented AI system can be characterised as an adaptive and cognitive system. The adaptive characteristic presents broader challenges beyond merely dealing with changes emanating from dynamic environments. To better describe this interaction, we use the word agent as a personification of the system from the user's perspective. From the human user perspective, the agent forms the communication interface between the system and the user. For human-agent interaction to be effective, the user must be able to specify their knowledge about the environment and decision-making context to the agent. 

The agent's knowledge base and reasoning processes constitute a shared model reflecting both the system’s and the user’s knowledge, reasoning and decision-making processes. This shared model serves as a basis for communication and interaction between the user and the system. The user may initially specify goals and objectives which are incomplete and vague. These will also change and evolve naturally over time as the agent adapts to the user and the user to the agent. Even though the agent may incorporate different inference algorithms, the agent must be able to follow reasoning patterns compatible with and understandable by the human user within their application context. In this way, the rationale and analysis of any decision recommendation can effectively be communicated and explained to the user. The human user can ask the question, “Why would I do this?” or “How did you arrive at this conclusion?”. As such, an agent must convince the user that the action it recommends is the best one to take after analysing all the information at hand. 

\subsection{Scientific knowledge discovery}\label{sec:sci_know_disc}
The One Health approach proposes an integrated, unifying approach to balance and optimize the health of people, animals and the environment. It will leverage scientific understanding and theories in the earth, biological and behavioural sciences. Physical, biological and social processes are constantly evolving and depend on dynamic micro-scale and meso-scale environments \cite{Moodley2012ArchitectureWeb}. Processes may vary significantly at different locations and typically change over time. Subsequently, theoretical positions differ significantly and may even be contradictory. An integral part of data analysis and scientific enquiry is the application of different theories to analyse incoming observations. Well-founded and established theories can be incorporated into modelling, simulation and decision-support tools. In contrast, new theories can be constantly adapted and evaluated against incoming observations to test their validity. Moreover, powerful Artificial Intelligence (AI) data mining and pattern analysis techniques can be continuously applied to data over time to capture new patterns, formulate new theories and refine existing theories. Ultimately the scientific community should publish dynamic and formal models of their theories and validation data online for immediate dissemination and reproducibility. This has the potential to accelerate and reduce the cost and effort for scientific discovery and lower the barrier for decision-makers to access the latest data analysis and simulation models to improve their planning and inform policy.

AI-driven knowledge discovery systems will have the capacity to pursue scientific research, collect measurements, find regularities, form hypotheses, and gather additional data to test them \cite{Langley2021AgentsDiscovery}. Like humans these continual learning systems \cite{Hadsell2020EmbracingNetworks} will learn incrementally and cumulatively. Key functions will include taxonomy formation, descriptive law induction and explanatory model construction \cite{Langley2021AgentsDiscovery}. These systems would engage in all of these scientific activities, each of which can involve detecting and responding to anomalous observations. While these systems may act autonomously \cite{Langley2021AgentsDiscovery}, a more compelling vision is for them to work in tandem with human researchers as a fully functional member of a research team. There are three fundamental functional areas where AI can contribute to new scientific understanding \cite{Krenn2022OnScientific}. First, AI can act as an instrument revealing properties of a physical system that are difficult or even impossible to probe. Humans then use these insights to formulate new scientific understanding. Second, AI can act as a source of inspiration for new concepts and ideas that are subsequently understood and generalized by human scientists. Third, AI acts as an agent to create new understanding. AI reaches new scientific insight and can transfer it to human researchers. 

\subsection{Knowledge based approaches}
Generating hypotheses and maintaining a consistent body of knowledge in science is a formidable task due to the vast number of hypotheses generated and maintained, and the complexity, non-monotonicity, uncertainty and unreliability of knowledge and data published \cite{Kitano2021Nobel}. Logic-based concept ontologies emanating from the Semantic Web community have been widely used to represent, and reason about concepts and relations in a domain \cite{Hitzler2020AReview}. Ontology languages like OWL are typically based on monotonic logic such as Description Logics and provide limited support to deal with uncertainty and non-monotonicity.  While they can effectively capture both syntax and semantics to enable communication, they are limited in terms of capturing and representing pragmatics, i.e., the context in which the knowledge is used \cite{Schoop2006Pragmatic, Singh2002ThePragmatic}. Ontologies can enable semantic interoperability, but not pragmatic interoperability between agents \cite{Neiva2016Towards}. Casanovas \cite{Casanovas2017TheRole} argue that social context is crucial and that meaning is only fully realised in actual situations assuming an interactional and dynamic notion of context. A further limitation is the lack of explicit support for representing decision-making processes. 

\subsubsection{Bayesian Networks and frameworks}
Bayesian Networks (BNs) are highly effective causal models that aim to capture human reasoning and decision-making in uncertain situations \cite{Korb2010Bayesian}. They support three different types of explanations \cite{Lacave2002AReview}. For the explanation of the evidence, an abduction inference process is typically used to obtain the most probable explanation (MPE) explanation of the evidence. The purpose of this kind of explanation is to offer a diagnosis for a set of observed anomalies.  Explanation of the model consists of displaying the information contained in the knowledge base and allows experts to easily navigate the knowledge. Explanation of reasoning provides justification for inferences made by the system. It allows users to understand and check the correctness of the reasoning process and the inferences made by the system. Compared to a rule-based expert system with IF-THEN rules, BNs are more compatible with present thinking about explanation \cite{Tesic2021Explanation}. However, it is neither clear how explanations in BNs can capture pragmatics, nor how to operationalise explanatory virtues in the context of BNs \cite{Tesic2021Explanation}. 

In philosophy, the conventional rational choice model of decision-making (based upon expected utility maximization), is widely discussed and used \cite{Briggs2014Normative}. Bayesian Decision Networks (BDN), which are based on expected utility maximisation, are widely used for representing decision-making processes. Fareh \cite{Fareh2019Modeling} discusses how Bayesian Networks (BN) can be combined with ontologies to represent incomplete knowledge and uncertainty to predict liver cancer. BDNs also incorporate uncertainty and can recommend decisions where there is incomplete knowledge. However, rational choice can be viewed as a prescriptive model, a way of specifying how individuals ought to behave, rather than how they actually behave. Human behaviour can also be impacted by emotion, mood, personality, needs and subjective well-being. Abaalkhail \cite{Abaalkhail2018Survey} reviews current ontologies for affective states (emotion and mood). These ontologies can be used to build decision-support systems that consider emotion and mood. In a recent review paper, Barthès \cite{Barthes2020Cognitive} explored decision-making mechanisms for ethical decision-making in cognitive agents. He highlights uncertainty and incomplete knowledge as key issues that must be taken into account. He provides a preliminary computational framework that takes into account different factors that influence human decision-making, including emotion and mood. This can be used as a foundation to design and analyse empathetic agents that are better able to support human decision-making.

\subsubsection{Explanation in AI systems}
Knowledge discovery, cognition, communication, decision making and explanation are integrally linked. In their recent papers, Chari et al \cite{Chari2020Directions, Chari2020Foundations} provide an excellent analysis of explanation in knowledge-based AI systems. They identify nine types of explanation \cite{Chari2020Directions} in AI systems and highlight the need for AI systems to present personalized, trustworthy, and context-aware explanations to users. Our view is that a 3D virtual twin can provide a visual explanation in both space and time and is grounded in a pseudo-reality. We consider this as a new type of explanation. This is linked to the notion of pragmatic and social world communication. A 3D virtual digital twin can provide a rich context for communicating predictions, theories and beliefs posed by the AI system. It provides the outputs and deliberations of the AI system to be rendered in a pseudo-reality for further analysis. 

\subsection{Continual learning}
Continual learning, also known as incremental learning or lifelong learning, is an increasingly relevant area of study that asks how artificial systems might learn sequentially, as biological systems do, from a continuous stream of data \cite{Hadsell2020EmbracingNetworks, Silver2013LifelongAlgorithms, vandeVen2022ThreeLearning}. Unlike, traditional Deep Neural Network (DNN) tasks that rely on fixed datasets and stationary environments, the problem of continual learning is defined by a sequential training protocol and by the features expected from the solution. In contrast to the common machine learning setting of a static dataset or environment, the continual learning setting explicitly focuses on non-stationary or changing environments.  Continual learning systems must deal with temporal generalisation, change detection and mechanisms for a model update, specifying and aligning expected features and loss functions, and appropriate validation and data partitioning methods. While some of these issues have been explored, e.g. prequential or walk-forward validation in time series forecasting \cite{Cerqueira2020EvaluatingMethods}, and more recently temporal generality in language models \cite{Lazaridou2021Mind}, this is still an emerging area with many open challenges.

Spatial Temporal Graph Neural Networks (ST-GNN) are a new wave of advanced DNN techniques, which have emerged recently to model and predict flow in complex systems \cite{Bui2022Spatial}. Typical characteristics are high frequency and noisy observations from multiple sensors with complex and often latent spatial and temporal dependencies. While the canonical application is for traffic flow in a city, these techniques have wider applications for modelling dynamic systems in general, for example, weather modelling \cite{Davidson2022ST-GNNs}. Weather prediction has a higher complexity than the traffic problem. Unlike traffic flow prediction which models a single variable, i.e. traffic speed at different points in a city, weather prediction involves many different weather variables at different temporal and spatial scales. A key feature of ST-GNNs is that they dynamically learn complex spatial-temporal dependencies inherent in the data and capture this in an adjacency matrix. These dependencies can be used as the foundation for constructing causal theories in the domain. We believe that ST-GNNs are at the frontier of learning-based approaches for knowledge discovery, predictive modelling and explanation in dynamic environments.

\subsection{Architectures for scientific knowledge discovery systems}
AI systems for scientific knowledge discovery are introduced in section \ref{sec:sci_know_disc}. In this section we explore some of the architectures for such systems. Systems that can contribute to new scientific understanding are certainly on the frontier of AI \cite{Langley2021AgentsDiscovery, Krenn2022OnScientific, Kitano2021Nobel}.  A compelling vision for the field is to design an “AI scientist” that works in tandem with human researchers as a fully functioning member of a research team. This overlaps with the notion of hybrid intelligence and augmented intelligence systems where the machine works in collaboration with and cooperatively with the human user \cite{Dellermann2019HybridIntelligence, Akata2020AResearch}. Hybrid intelligence systems for knowledge discovery have been explored recently by Gil in her position paper on “Thoughtful Artificial Intelligence Systems” \cite{GilThoughtfulDiscovery} where she proposes seven principles for such systems and sets an agenda for research in this area. 

\subsubsection{Ontology driven systems}
One of the first examples of knowledge discovery systems is the Robot Scientist \cite{King2009TheAdam}. The Robot Scientist is a semantic architecture that incorporates the LABORS ontology. 
More recent efforts are DISK \cite{Gil2016AutomatedRepositories} and HELO \cite{Soldatova2013RepresentationKnowledge}. In the HELO study \cite{Soldatova2013RepresentationKnowledge} the representation of uncertainty is identified as a key limitation of previous ontology-driven knowledge discovery systems. HELO explicitly targets this limitation and Bayesian probability is incorporated into the HELO ontology to represent the current belief or uncertainty of different research statements or hypotheses. The uncertainty of research statements is updated as new evidence arises, forming the basis for decision-making during the research process. This aligns with our previous work where we found that a knowledge engineering approach that uses ontologies alone has substantial limitations in real-world applications. Based on our experience in four studies in South Africa across diverse domains and user communities, i.e. earth observation \cite{Moodley2012ArchitectureWeb}, health \cite{Ogundele2016AnOntology}, biodiversity \cite{Coetzer2017AKnowledge}, and finance \cite{Drake2022INVEST:JSE}, we found that an approach that combines both ontologies and Bayesian decision networks to be highly effective to deal with these challenges. While ontologies are ideal for representing and structuring domain knowledge, Bayesian decision networks (BDN) provide explicit support for reasoning with uncertainty, and for capturing and reasoning about decision-making processes.

\subsubsection{Emerging architectures based on Philosophy of Science}
Understanding scientific method is essential for designing knowledge discovery systems. Various theories of scientific method have been proposed in Philosophy of Science, for example, the inductive theory and the hypothetico-deductive theory of methods. The scientific Abductive Theory of Method (ATOM) \cite{Haig2005AnMethod, Haig2018AnMethod} is a recent theory that is more encompassing and detailed and is compatible with current thinking around knowledge discovery in the AI community \cite{Langley2021AgentsDiscovery,Krenn2022OnScientific, Kitano2021Nobel} (described above). ATOM consists of two overarching processes, namely i) phenomenon detection where novel or anomalous patterns are detected in data, and ii) theory construction where plausible theories are generated and evaluated to explain detected phenomena. In recent work \cite{Wanyana2021AnEvolution}, we explored a generic architecture for a knowledge discovery agent based on ATOM. There are many advantages for grounding the architecture in ATOM. It will more closely align with the process that researchers actually use making it more human-centred.  It provides an abstract architecture that incorporates both data-driven and knowledge-based AI approaches described above. It also provides an entry point to the large body of recent work emanating from the Philosophy of Science community on theory construction. Haig suggests the use of analogical reasoning for theory construction and for assessing explanations \cite{Haig2005AnMethod}. Confirmation based on analogical inference using a Bayesian framework is currently being explored in the Philosophy of Science community \cite{Dardashti2019HawkingAnalysis, Feldbacher-Escamilla2020ConfirmationJeffrey}. Analogical inference is related to the notion of transfer learning in the machine learning community. This is particularly important in low-resource settings where large clinical data sets may not be available. Machine learning models may be training initially on large data sets from diverse populations in the developed world and then fine tuned on smaller data sets in low resource African countries. An initial exploration of transfer learning to detect heart disease in diverse populations from ECG signals can be found in \cite{Aarons2022AGeneralizable}. 

\section{An architecture for an AI-driven digital twin system}
\begin{figure}
  \centering
  \includegraphics[width=\linewidth]{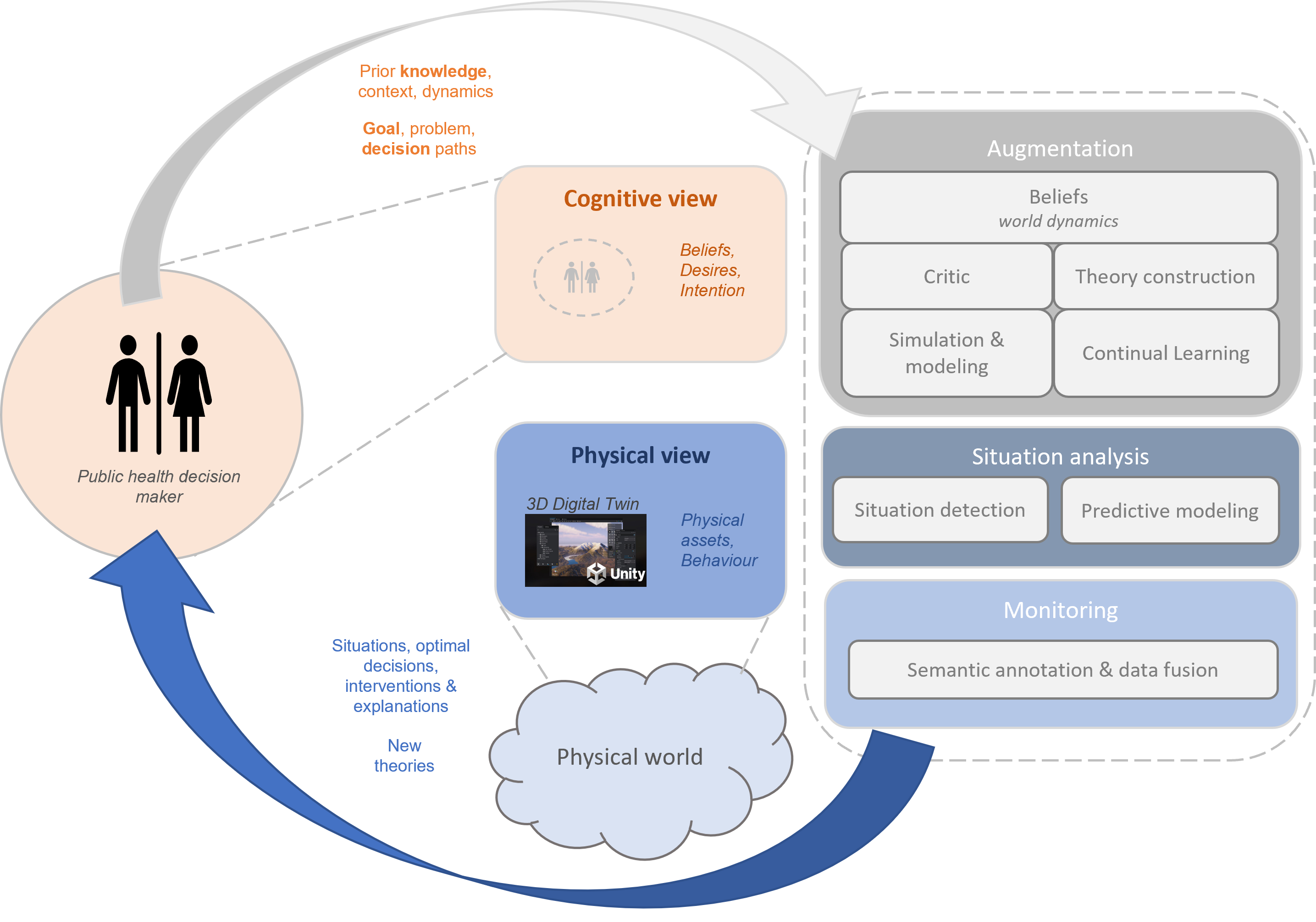}
  \caption{An architecture for an AI-driven 3D digital twin}
  \label{fig:ai-dti-architecture}
\end{figure}

We propose an initial abstract architecture for an augmented AI system in figure \ref{fig:ai-dti-architecture}. The architecture draws from and extends the ideas presented in Wanyana and Moodley's KDE Agent architecture \cite{Wanyana2021AnEvolution}, Russell and Norvig's learning agent \cite{Russell2009ArtificialApproach} and Ricci et al's WoDT \cite{Ricci2022WebTwins}. The AI system augments the 3D digital twin and uses it as a key platform for its simulations, for knowledge discovery and ultimately to augment the human user, in this case, the public health decision maker.  

\subsection{System overview}
The system will continuously interrogate vast quantities of heterogeneous observational data, automatically generate and maintain internal models to explain evolving phenomena, evaluate these models through simulation in the DT and in cooperation with the user, discover a new understanding of evolving phenomena and provide possible courses of actions to achieve the goals set by the human user. 

The system maintains two views of the world which connect it to the user and the physical world. The Cognitive view maintains a model of the user, i.e. the beliefs, goals and decision paths of the public health decision-maker. The Physical view encompasses the 3D digital world, which is used to model a pseudo-reality of the physical world, including properties of physical entities, and interactions and behaviours of individuals and populations. 

Adaptation and cognition are two essential functional areas of the system. The adaptive characteristic is concerned with rapid learning and model updates when the world changes, communicating new information about the environment to the user and adjusting to changes in the user’s decision-making goals and preferences as the user learns and adjusts to the system. The cognitive characteristic involves the use of representing prior knowledge, world dynamics and appropriate reasoning mechanisms for sense-making and belief update and revision that is aligned to human reasoning. The cognitive aspect facilitates the overall interactions with the user and provides explicit support for providing explanations and allows the user to interrogate courses of action recommended by the system.

The abstract architecture consists of three layers, i.e. the Monitoring Layer, the Situation Analysis Layer and the Augmentation Layer

\subsection{Monitoring Layer}
At the Monitoring Layer, ontologies will be used for semantic data annotation and data fusion. Data will include earth observation data, clinical data, population data, behavioural data and clinical data. Existing semantic architectures from our previous work in earth observation \cite{Moodley2012ArchitectureWeb} and biodiversity \cite{Coetzer2017AKnowledge} will be leveraged for semantic mediation and dynamic data fusion. 

The Sensor Web Agent Platform (SWAP) architecture \cite{Moodley2012ArchitectureWeb} proposes a three-layered open distributed system architecture and a conceptual knowledge representation and reasoning framework that integrates ontologies and Bayesian networks for developing distributed Sensor Web applications. The SWAP semantic architecture provides modular ontologies for reasoning about space, time, theme (the entity and properties being observed) and uncertainty and supports semantic annotation and dynamic data fusion from heterogeneous sensor data.  Coetzer et al \cite{Coetzer2017AKnowledge} explored a semantic architecture for a knowledge-based system that uses expert knowledge to generate ecological interaction networks from distributed and heterogeneous natural history occurrence data. The system builds on the SWAP architecture and introduces a biodiversity mapping ontology that supports semantic annotation and fusion of heterogeneous data from three biodiversity databases hosted by different museums across South Africa.

The 3D digital twin provides a pseudo-reality of the physical world. Observational data stored in databases in existing systems together with real-time sensor observations will be semantically annotated, fused and rendered in the 3D digital twin. The fused observations are  contextualised within the 3D world and must be harmonised and logically consistent with other observations, entities and processes that are unfolding in the 3D world. The 3D world thus provides context for and guides data fusion and is integral to achieving pragmatic interoperability.

\subsection{Situation Analysis}
This layer consists of two components. Situation detection can use ontologies, semantic rules, fuzzy rules and Bayesian networks for detecting current situations. Semantic technologies are increasingly being used for situation analysis in personal health monitoring systems that incorporate wearable sensors \cite{Nzomo2023Semantic}. Bayesian networks can be used for situation analysis and root cause analysis to support clinical diagnosis \cite{Wanyana2022Combining}. They can be supplemented by machine learning, e.g. ST-GNNs for flow prediction for early detection of adverse events. 

Ontologies and Bayesian networks can be combined for behavioural modelling and risk analysis, based on an approach we used in our previous work on analysing the risk of non-adherence behaviour for medical treatment for tuberculosis patients \cite{Ogundele2016AnOntology}. 

\subsection{Augmentation Layer}
The Augmentation Layer has three functional components. Similarly to the CDT, the system augments the DT and in turn augments the user. This tridirectional augmentation is a crucial aspect of the architecture. Each environment evolves separately, but feedback loops between the user, the DT and the AI system allow for synchronisation and alignment. 

The system maintains the current beliefs in a knowledge base. It uses continual learning, e.g. STGNNs, for discovering novel spatial-temporal patterns or new phenomena. This triggers the theory construction module, which attempts to find explanations for the phenomena. If there is no suitable explanation, it uses 3D simulation in the DT to better understand the nature and possible root causes of the novel phenomena. These possible explanations form new unconfirmed theories that must be further explored. The latest theories are exposed to the user in the DT for further analysis and input. The user can prioritise theories or propose alternate theories not discovered by the system. The critic module, similar to the critic module in Russell \& Norvigs Learning Agent \cite{Russell2009ArtificialApproach}, provides internal evaluation and analysis of new theories. It prioritises theories aligned with the user's goals and uses the DT to evaluate theories by integrating knowledge from different contexts and perspectives. 

The key notion is explicitly modelling the human decision maker's mental models, i.e. the Cognitive view. We see the digital twin as capturing physical notions, but the specification of beliefs, goals, decisions, and knowledge are abstract notions which we separate from the physical representation. The physical representation (Physical view) not only models properties of physical entities but also interactions and behaviours of individual people and populations. 

The architecture for an  AI-driven digital twin system shown in Figure \ref{fig:ai-dti-architecture} aligns with the public health decision-making process in a PHEOC, described in section \ref{sec:PHEOC}. An AI-driven digital twin can represent and support different activities and, in addition, provide enhanced support for predictive analytics, knowledge discovery and interactive decision-making.

\section{Discussion and conclusion}
This paper describes an ambitious and bold vision to harness the potential of artificial intelligence and digital twins to re-imagine public health in African countries. A future digital platform for One Health will have to overcome both social and technical challenges. An open and distributed One Health platform must mediate between many heterogeneous systems which manage population, clinical, vital statistics, geospatial and environmental data across different government departments at the facility, city, district, state and country levels.

The application of Digital Twins to public health shows much promise in LMICs. An opportunity exists to use a 3D digital twin to mediate and contextualise different data sources, models and systems from a public health perspective. This could include dynamic fusion and harmonisation of multiple information sources and models in an open distributed environment with different contributors and users. Equipped with AI, the system could also provide predictive modelling and risk profiling to mitigate disease outbreaks and support early warning systems for adverse health conditions. The digital twin provides a powerful visual analysis and simulation tool for investigating and explaining the root causes of events of interest in public health. However, it will be important to understand the ethical implications of these approaches.

To truly unleash the power of these emerging technologies would require a continental-scale effort and buy-in across multiple African countries, ministries of health, international donors, universities and non-governmental organisations. Establishing standards and platforms and an ecosystem to support this vision will indeed be challenging, especially in low-resource environments. The structure of the ecosystem must balance long-term sustainability, provide a clear vision, and must be resilient to changing technologies and organisational changes in the health system. Based on our previous experience in the public health sector our view is that a middle-out strategy \cite{Mudaly2013ArchitecturalCountries} should be considered for structuring such an ecosystem. In such a middle-out approach, a central authority, such as the national government, manages and provides leadership in setting standards, the platform architecture and the overall direction while still allowing autonomy for participating organisations to implement and manage their own systems.

\bibliography{references.bib}

\begin{thebibliography}{75}
\expandafter\ifx\csname natexlab\endcsname\relax\def\natexlab#1{#1}\fi
\providecommand{\url}[1]{\texttt{#1}}
\providecommand{\href}[2]{#2}
\providecommand{\path}[1]{#1}
\providecommand{\DOIprefix}{doi:}
\providecommand{\ArXivprefix}{arXiv:}
\providecommand{\URLprefix}{URL: }
\providecommand{\Pubmedprefix}{pmid:}
\providecommand{\doi}[1]{\href{http://dx.doi.org/#1}{\path{#1}}}
\providecommand{\Pubmed}[1]{\href{pmid:#1}{\path{#1}}}
\providecommand{\bibinfo}[2]{#2}
\ifx\xfnm\relax \def\xfnm[#1]{\unskip,\space#1}\fi
\bibitem[{{WHO}(1978)}]{WHO1978Primary}
\bibinfo{author}{{WHO}}, \bibinfo{title}{{Declaration of Alma-Ata}},
  \bibinfo{type}{Technical Report}, \bibinfo{year}{1978}. \URLprefix
  \url{https://www.who.int/publications/i/item/WHO-EURO-1978-3938-43697-61471}.
\bibitem[{{WHO}(2021)}]{WHO2021World}
\bibinfo{author}{{WHO}},
\newblock \bibinfo{title}{{World Health Statistics 2021}}
  (\bibinfo{year}{2021}). \URLprefix
  \url{https://apps.who.int/iris/bitstream/handle/10665/342703/9789240027053-eng.pdf}.
\bibitem[{Coates et~al.(2021)Coates, Ezzati, Aguilar, Kwan, Vigo, Mocumbi,
  Becker, Makani, Hyder, Jain, Cristina~Stefan, Gupta, Marx, and
  Bukhman}]{Coates2021Burden}
\bibinfo{author}{M.~M. Coates}, \bibinfo{author}{M.~Ezzati},
  \bibinfo{author}{G.~R. Aguilar}, \bibinfo{author}{G.~F. Kwan},
  \bibinfo{author}{D.~Vigo}, \bibinfo{author}{A.~O. Mocumbi},
  \bibinfo{author}{A.~E. Becker}, \bibinfo{author}{J.~Makani},
  \bibinfo{author}{A.~A. Hyder}, \bibinfo{author}{Y.~Jain},
  \bibinfo{author}{D.~Cristina~Stefan}, \bibinfo{author}{N.~Gupta},
  \bibinfo{author}{A.~Marx}, \bibinfo{author}{G.~Bukhman},
\newblock \bibinfo{title}{{Burden of disease among the world’s poorest
  billion people: An expert-informed secondary analysis of Global Burden of
  Disease estimates}},
\newblock \bibinfo{journal}{PLoS ONE} \bibinfo{volume}{16}
  (\bibinfo{year}{2021}). \DOIprefix\doi{10.1371/journal.pone.0253073}.
\bibitem[{Natifu(2020)}]{Nalifu2020Climate}
\bibinfo{author}{B.~Natifu}, \bibinfo{title}{{Climate Change and Health in
  Sub-Saharan Africa: The Case of Uganda}}, \bibinfo{type}{Technical Report},
  \bibinfo{year}{2020}. \URLprefix
  \url{https://www.cif.org/sites/cif_enc/files/knowledge-documents/final_chasa_report_19may2020.pdf}.
\bibitem[{Ciecierski-Holmes et~al.(2022)Ciecierski-Holmes, Singh, Axt, Brenner,
  and Barteit}]{Ciecierski-Holmes2022AI}
\bibinfo{author}{T.~Ciecierski-Holmes}, \bibinfo{author}{R.~Singh},
  \bibinfo{author}{M.~Axt}, \bibinfo{author}{S.~Brenner},
  \bibinfo{author}{S.~Barteit}, \bibinfo{title}{{Artificial intelligence for
  strengthening healthcare systems in low- and middle-income countries: a
  systematic scoping review}}, \bibinfo{year}{2022}. \URLprefix
  \url{https://www.nature.com/articles/s41746-022-00700-y}.
  \DOIprefix\doi{10.1038/s41746-022-00700-y}.
\bibitem[{{The Lancet}(2023)}]{TheLancet2023OneHealth}
\bibinfo{author}{{The Lancet}}, \bibinfo{title}{{One Health: a call for
  ecological equity}}, \bibinfo{year}{2023}.
  \DOIprefix\doi{10.1016/S0140-6736(23)00090-9}.
\bibitem[{Worsley-Tonks et~al.(2022)Worsley-Tonks, Bender, Deem, Ferguson,
  F{\`{e}}vre, Martins, Muloi, Murray, Mutinda, Ogada, Omondi, Prasad, Wild,
  Zimmerman, and Hassell}]{Worsley-Tonks2022Strengthening}
\bibinfo{author}{K.~E. Worsley-Tonks}, \bibinfo{author}{J.~B. Bender},
  \bibinfo{author}{S.~L. Deem}, \bibinfo{author}{A.~W. Ferguson},
  \bibinfo{author}{E.~M. F{\`{e}}vre}, \bibinfo{author}{D.~J. Martins},
  \bibinfo{author}{D.~M. Muloi}, \bibinfo{author}{S.~Murray},
  \bibinfo{author}{M.~Mutinda}, \bibinfo{author}{D.~Ogada},
  \bibinfo{author}{G.~P. Omondi}, \bibinfo{author}{S.~Prasad},
  \bibinfo{author}{H.~Wild}, \bibinfo{author}{D.~M. Zimmerman},
  \bibinfo{author}{J.~M. Hassell}, \bibinfo{title}{{Strengthening global health
  security by improving disease surveillance in remote rural areas of
  low-income and middle-income countries}}, \bibinfo{year}{2022}. \URLprefix
  \url{https://www.thelancet.com/journals/langlo/article/PIIS2214-109X(22)00031-6/fulltext}.
  \DOIprefix\doi{10.1016/S2214-109X(22)00031-6}.
\bibitem[{One(2022)}]{OneHealth2022}
\bibinfo{title}{{Working together for the health of humans, animals, plants and
  the environment. One Health Joint Plan of Action (2022–2026)}},
\newblock \bibinfo{journal}{Bulletin de l'OIE} \bibinfo{volume}{2022}
  (\bibinfo{year}{2022}) \bibinfo{pages}{18--19}.
  \DOIprefix\doi{10.20506/bull.2022.2.3324}.
\bibitem[{{World Health Organisation}(2015)}]{WHO2015Framework}
\bibinfo{author}{{World Health Organisation}}, \bibinfo{title}{{Framework for a
  Public Health Emergency Operations Centre}}, \bibinfo{publisher}{World Health
  Organisation}, \bibinfo{year}{2015}. \URLprefix
  \url{https://www.who.int/publications/i/item/framework-for-a-public-health-emergency-operations-centre}.
\bibitem[{{WHO}(2021)}]{WHO2021Handbook}
\bibinfo{author}{{WHO}}, \bibinfo{title}{{Handbook for Public Health Emergency
  Operations Center Operations and Management.}}, \bibinfo{year}{2021}.
  \URLprefix
  \url{https://africacdc.org/download/handbook-for-public-health-emergency-operations-center-operations-and-management/}.
\bibitem[{{Ethiopia Public Health Institute (EPHI)}(2022)}]{EPHI2022National}
\bibinfo{author}{{Ethiopia Public Health Institute (EPHI)}},
  \bibinfo{title}{{National Public Health Emergency Operation Center
  Handbook}}, \bibinfo{type}{Technical Report}, \bibinfo{year}{2022}.
  \URLprefix
  \url{https://ephi.gov.et/wp-content/uploads/2022/07/EPHI_cPHEM_EWISMD_PHEOC_Handbook_V1.pdf}.
\bibitem[{{UNAIDS}(2020)}]{UNAIDS2020}
\bibinfo{author}{{UNAIDS}}, \bibinfo{title}{{Fast-Track Targets}},
  \bibinfo{type}{Technical Report}, \bibinfo{year}{2020}. \URLprefix
  \url{https://www.unaids.org/sites/default/files/media_asset/201506_JC2743_Understanding_FastTrack_en.pdf}.
\bibitem[{{Pepfar}(2022)}]{PEPFAR2022}
\bibinfo{author}{{Pepfar}}, \bibinfo{title}{{Reimagining PEPFAR's Strategic
  Direction}}, \bibinfo{type}{Technical Report}, \bibinfo{year}{2022}.
  \URLprefix
  \url{https://www.state.gov/wp-content/uploads/2022/09/PEPFAR-Strategic-Direction_FINAL.pdf}.
\bibitem[{Rice et~al.(2018)Rice, Boulle, Baral, Egger, Mee, Fearon, Reniers,
  Todd, Schwarcz, Weir, Rutherford, and Hargreaves}]{Rice2018Strengthening}
\bibinfo{author}{B.~Rice}, \bibinfo{author}{A.~Boulle},
  \bibinfo{author}{S.~Baral}, \bibinfo{author}{M.~Egger},
  \bibinfo{author}{P.~Mee}, \bibinfo{author}{E.~Fearon},
  \bibinfo{author}{G.~Reniers}, \bibinfo{author}{J.~Todd},
  \bibinfo{author}{S.~Schwarcz}, \bibinfo{author}{S.~Weir},
  \bibinfo{author}{G.~Rutherford}, \bibinfo{author}{J.~Hargreaves},
\newblock \bibinfo{title}{{Strengthening routine data systems to track the HIV
  epidemic and guide the response in Sub-Saharan Africa}},
\newblock \bibinfo{journal}{JMIR Public Health and Surveillance}
  \bibinfo{volume}{4} (\bibinfo{year}{2018}).
  \DOIprefix\doi{10.2196/publichealth.9344}.
\bibitem[{Moodley et~al.(2018)Moodley, Pillay, and
  Seebregts}]{Moodley2018Establishing}
\bibinfo{author}{D.~Moodley}, \bibinfo{author}{A.~Pillay},
  \bibinfo{author}{J.~Seebregts},
\newblock \bibinfo{title}{{Establishing a health informatics research
  laboratory in South Africa}},
\newblock in: \bibinfo{booktitle}{Digital Re-imagination Colloquium 2018:
  Preparing South Africa for a Digital Future through e-Skills},
  \bibinfo{publisher}{NEMISA}, \bibinfo{year}{2018}, pp.
  \bibinfo{pages}{16--24}. \URLprefix
  \url{https://www.cair.org.za/sites/default/files/2020-02/HeAL-NEMISA-2018.pdf}.
\bibitem[{Kamel~Boulos and Zhang(2021)}]{KamelBoulos2021DigitalHealth}
\bibinfo{author}{M.~N. Kamel~Boulos}, \bibinfo{author}{P.~Zhang},
  \bibinfo{title}{{Digital twins: From personalised medicine to precision
  public health}}, \bibinfo{year}{2021}. \DOIprefix\doi{10.3390/jpm11080745}.
\bibitem[{Ahmadi-Assalemi et~al.(2020)Ahmadi-Assalemi, Al-Khateeb, Maple,
  Epiphaniou, Alhaboby, Alkaabi, and Alhaboby}]{Assalemi2020Digital}
\bibinfo{author}{G.~Ahmadi-Assalemi}, \bibinfo{author}{H.~Al-Khateeb},
  \bibinfo{author}{C.~Maple}, \bibinfo{author}{G.~Epiphaniou},
  \bibinfo{author}{Z.~A. Alhaboby}, \bibinfo{author}{S.~Alkaabi},
  \bibinfo{author}{D.~Alhaboby},
\newblock \bibinfo{title}{{Digital twins for precision healthcare}},
\newblock in: \bibinfo{booktitle}{Advanced Sciences and Technologies for
  Security Applications}, \bibinfo{year}{2020}.
  \DOIprefix\doi{10.1007/978-3-030-35746-7{\_}8}.
\bibitem[{Coorey et~al.(2021)Coorey, Figtree, Fletcher, and
  Redfern}]{Coorey2021TheHealth}
\bibinfo{author}{G.~Coorey}, \bibinfo{author}{G.~A. Figtree},
  \bibinfo{author}{D.~F. Fletcher}, \bibinfo{author}{J.~Redfern},
  \bibinfo{title}{{The health digital twin: advancing precision cardiovascular
  medicine}}, \bibinfo{year}{2021}. \DOIprefix\doi{10.1038/s41569-021-00630-4}.
\bibitem[{Fuller et~al.(2020)Fuller, Fan, Day, and Barlow}]{Fuller2020Digital}
\bibinfo{author}{A.~Fuller}, \bibinfo{author}{Z.~Fan},
  \bibinfo{author}{C.~Day}, \bibinfo{author}{C.~Barlow},
\newblock \bibinfo{title}{{Digital Twin: Enabling Technologies, Challenges and
  Open Research}},
\newblock \bibinfo{journal}{IEEE Access} \bibinfo{volume}{8}
  (\bibinfo{year}{2020}) \bibinfo{pages}{108952--108971}.
  \DOIprefix\doi{10.1109/ACCESS.2020.2998358}.
\bibitem[{Abburu et~al.(2020)Abburu, Berre, Jacoby, Roman, Stojanovic, and
  Stojanovic}]{Abburu2020COGNITWIN}
\bibinfo{author}{S.~Abburu}, \bibinfo{author}{A.~J. Berre},
  \bibinfo{author}{M.~Jacoby}, \bibinfo{author}{D.~Roman},
  \bibinfo{author}{L.~Stojanovic}, \bibinfo{author}{N.~Stojanovic},
\newblock \bibinfo{title}{{COGNITWIN - Hybrid and Cognitive Digital Twins for
  the Process Industry}},
\newblock in: \bibinfo{booktitle}{Proceedings - 2020 IEEE International
  Conference on Engineering, Technology and Innovation, ICE/ITMC 2020},
  \bibinfo{year}{2020}. \DOIprefix\doi{10.1109/ICE/ITMC49519.2020.9198403}.
\bibitem[{Zheng et~al.(2022)Zheng, Lu, and Kiritsis}]{Zheng2020TheEmergence}
\bibinfo{author}{X.~Zheng}, \bibinfo{author}{J.~Lu},
  \bibinfo{author}{D.~Kiritsis},
\newblock \bibinfo{title}{{The emergence of cognitive digital twin: vision,
  challenges and opportunities}},
\newblock \bibinfo{journal}{International Journal of Production Research}
  \bibinfo{volume}{60} (\bibinfo{year}{2022}) \bibinfo{pages}{7610--7632}.
  \DOIprefix\doi{10.1080/00207543.2021.2014591}.
\bibitem[{Ricci et~al.(2022{\natexlab{a}})Ricci, Croatti, Mariani, Montagna,
  and Picone}]{Ricci2022WebTwins}
\bibinfo{author}{A.~Ricci}, \bibinfo{author}{A.~Croatti},
  \bibinfo{author}{S.~Mariani}, \bibinfo{author}{S.~Montagna},
  \bibinfo{author}{M.~Picone},
\newblock \bibinfo{title}{{Web of Digital Twins}},
\newblock \bibinfo{journal}{ACM Transactions on Internet Technology}
  \bibinfo{volume}{22} (\bibinfo{year}{2022}{\natexlab{a}})
  \bibinfo{pages}{1--30}. \DOIprefix\doi{10.1145/3507909}.
\bibitem[{Ricci et~al.(2022{\natexlab{b}})Ricci, Croatti, and
  Montagna}]{Ricci2021Pervasive}
\bibinfo{author}{A.~Ricci}, \bibinfo{author}{A.~Croatti},
  \bibinfo{author}{S.~Montagna},
\newblock \bibinfo{title}{{Pervasive and Connected Digital Twins - A Vision for
  Digital Health}},
\newblock \bibinfo{journal}{IEEE Internet Computing} \bibinfo{volume}{26}
  (\bibinfo{year}{2022}{\natexlab{b}}) \bibinfo{pages}{26--32}.
  \DOIprefix\doi{10.1109/MIC.2021.3052039}.
\bibitem[{Hassani et~al.(2022)Hassani, Huang, and
  MacFeely}]{Hassani2022Impactful}
\bibinfo{author}{H.~Hassani}, \bibinfo{author}{X.~Huang},
  \bibinfo{author}{S.~MacFeely},
\newblock \bibinfo{title}{{Impactful Digital Twin in the Healthcare
  Revolution}},
\newblock \bibinfo{journal}{Big Data and Cognitive Computing}
  \bibinfo{volume}{6} (\bibinfo{year}{2022}).
  \DOIprefix\doi{10.3390/bdcc6030083}.
\bibitem[{Chen et~al.(2022)Chen, AlNajem, and Shorfuzzaman}]{Chen2022Digital}
\bibinfo{author}{D.~Chen}, \bibinfo{author}{N.~A. AlNajem},
  \bibinfo{author}{M.~Shorfuzzaman},
\newblock \bibinfo{title}{{Digital twins to fight against COVID-19 pandemic}},
\newblock \bibinfo{journal}{Internet of Things and Cyber-Physical Systems}
  \bibinfo{volume}{2} (\bibinfo{year}{2022}) \bibinfo{pages}{70--81}.
  \DOIprefix\doi{10.1016/j.iotcps.2022.05.003}.
\bibitem[{Rocha et~al.(2019)Rocha, Dias, Santinha, Rodrigues, Queir{\'{o}}s,
  and Rodrigues}]{Rocha2019Smart}
\bibinfo{author}{N.~P. Rocha}, \bibinfo{author}{A.~Dias},
  \bibinfo{author}{G.~Santinha}, \bibinfo{author}{M.~Rodrigues},
  \bibinfo{author}{A.~Queir{\'{o}}s}, \bibinfo{author}{C.~Rodrigues},
\newblock \bibinfo{title}{{Smart Cities and Public Health: A Systematic
  Review}},
\newblock in: \bibinfo{booktitle}{Procedia Computer Science}, volume
  \bibinfo{volume}{164}, \bibinfo{publisher}{Elsevier B.V.},
  \bibinfo{year}{2019}, pp. \bibinfo{pages}{516--523}.
  \DOIprefix\doi{10.1016/j.procs.2019.12.214}.
\bibitem[{Khan et~al.(2022)Khan, Milne-Ives, Meinert, Iyawa, Jones, and
  Josephraj}]{Khan2022AScoping}
\bibinfo{author}{A.~Khan}, \bibinfo{author}{M.~Milne-Ives},
  \bibinfo{author}{E.~Meinert}, \bibinfo{author}{G.~E. Iyawa},
  \bibinfo{author}{R.~B. Jones}, \bibinfo{author}{A.~N. Josephraj},
\newblock \bibinfo{title}{{A Scoping Review of Digital Twins in the Context of
  the Covid-19 Pandemic}},
\newblock \bibinfo{journal}{Biomedical Engineering and Computational Biology}
  \bibinfo{volume}{13} (\bibinfo{year}{2022}).
  \DOIprefix\doi{10.1177/11795972221102115}.
\bibitem[{Elkefi and Asan(2022)}]{Elkefi2022Digital}
\bibinfo{author}{S.~Elkefi}, \bibinfo{author}{O.~Asan},
  \bibinfo{title}{{Digital Twins for Managing Health Care Systems: Rapid
  Literature Review}}, \bibinfo{year}{2022}. \DOIprefix\doi{10.2196/37641}.
\bibitem[{Zheng et~al.(2017)Zheng, Liu, Ren, Ma, Chen, Yu, Xue, Chen, and
  Wang}]{Zheng2017Hybrid}
\bibinfo{author}{N.~n. Zheng}, \bibinfo{author}{Z.~y. Liu},
  \bibinfo{author}{P.~j. Ren}, \bibinfo{author}{Y.~q. Ma},
  \bibinfo{author}{S.~t. Chen}, \bibinfo{author}{S.~y. Yu},
  \bibinfo{author}{J.~r. Xue}, \bibinfo{author}{B.~d. Chen},
  \bibinfo{author}{F.~y. Wang}, \bibinfo{title}{{Hybrid-augmented intelligence:
  collaboration and cognition}}, \bibinfo{year}{2017}.
  \DOIprefix\doi{10.1631/FITEE.1700053}.
\bibitem[{Yau et~al.(2021)Yau, Lee, Chong, Ling, Syed, Wu, and
  Goh}]{Yau2021Augmented}
\bibinfo{author}{K.~L.~A. Yau}, \bibinfo{author}{H.~J. Lee},
  \bibinfo{author}{Y.~W. Chong}, \bibinfo{author}{M.~H. Ling},
  \bibinfo{author}{A.~R. Syed}, \bibinfo{author}{C.~Wu}, \bibinfo{author}{H.~G.
  Goh},
\newblock \bibinfo{title}{{Augmented Intelligence: Surveys of Literature and
  Expert Opinion to Understand Relations between Human Intelligence and
  Artificial Intelligence}},
\newblock \bibinfo{journal}{IEEE Access} \bibinfo{volume}{9}
  (\bibinfo{year}{2021}) \bibinfo{pages}{136744--136761}.
  \DOIprefix\doi{10.1109/ACCESS.2021.3115494}.
\bibitem[{Akata et~al.(2020)Akata, Balliet, De~Rijke, Dignum, Dignum, Eiben,
  Fokkens, Grossi, Hindriks, Hoos, Hung, Jonker, Monz, Neerincx, Oliehoek,
  Prakken, Schlobach, Van Der~Gaag, Van~Harmelen, Van~Hoof, Van~Riemsdijk,
  Van~Wynsberghe, Verbrugge, Verheij, Vossen, and Welling}]{Akata2020AResearch}
\bibinfo{author}{Z.~Akata}, \bibinfo{author}{D.~Balliet},
  \bibinfo{author}{M.~De~Rijke}, \bibinfo{author}{F.~Dignum},
  \bibinfo{author}{V.~Dignum}, \bibinfo{author}{G.~Eiben},
  \bibinfo{author}{A.~Fokkens}, \bibinfo{author}{D.~Grossi},
  \bibinfo{author}{K.~Hindriks}, \bibinfo{author}{H.~Hoos},
  \bibinfo{author}{H.~Hung}, \bibinfo{author}{C.~Jonker},
  \bibinfo{author}{C.~Monz}, \bibinfo{author}{M.~Neerincx},
  \bibinfo{author}{F.~Oliehoek}, \bibinfo{author}{H.~Prakken},
  \bibinfo{author}{S.~Schlobach}, \bibinfo{author}{L.~Van Der~Gaag},
  \bibinfo{author}{F.~Van~Harmelen}, \bibinfo{author}{H.~Van~Hoof},
  \bibinfo{author}{B.~Van~Riemsdijk}, \bibinfo{author}{A.~Van~Wynsberghe},
  \bibinfo{author}{R.~Verbrugge}, \bibinfo{author}{B.~Verheij},
  \bibinfo{author}{P.~Vossen}, \bibinfo{author}{M.~Welling},
\newblock \bibinfo{title}{{A Research Agenda for Hybrid Intelligence:
  Augmenting Human Intellect with Collaborative, Adaptive, Responsible, and
  Explainable Artificial Intelligence}},
\newblock \bibinfo{journal}{Computer} \bibinfo{volume}{53}
  (\bibinfo{year}{2020}) \bibinfo{pages}{18--28}.
  \DOIprefix\doi{10.1109/MC.2020.2996587}.
\bibitem[{Zhang et~al.(2021)Zhang, Ning, Liu, Jin, and Piuri}]{Zhang2021Hybrid}
\bibinfo{author}{W.~Zhang}, \bibinfo{author}{H.~Ning},
  \bibinfo{author}{L.~Liu}, \bibinfo{author}{Q.~Jin},
  \bibinfo{author}{V.~Piuri}, \bibinfo{title}{{Guest Editorial: Special Issue
  on Hybrid Human-Artificial Intelligence for Social Computing}},
  \bibinfo{year}{2021}. \DOIprefix\doi{10.1109/TCSS.2021.3049702}.
\bibitem[{Moodley et~al.(2012)Moodley, Simonis, and
  Tapamo}]{Moodley2012ArchitectureWeb}
\bibinfo{author}{D.~Moodley}, \bibinfo{author}{I.~Simonis},
  \bibinfo{author}{J.~Tapamo},
\newblock \bibinfo{title}{{Architecture for managing knowledge and system
  dynamism in the worldwide sensor web}},
\newblock \bibinfo{journal}{International Journal on Semantic Web and
  Information Systems} \bibinfo{volume}{8} (\bibinfo{year}{2012}).
  \DOIprefix\doi{10.4018/jswis.2012010104}.
\bibitem[{Langley(2021)}]{Langley2021AgentsDiscovery}
\bibinfo{author}{P.~Langley},
\newblock \bibinfo{title}{{Agents of exploration and discovery}}
  (\bibinfo{year}{2021}). \DOIprefix\doi{10.1609/aaai.12021}.
\bibitem[{Hadsell et~al.(2020)Hadsell, Rao, Rusu, and
  Pascanu}]{Hadsell2020EmbracingNetworks}
\bibinfo{author}{R.~Hadsell}, \bibinfo{author}{D.~Rao}, \bibinfo{author}{A.~A.
  Rusu}, \bibinfo{author}{R.~Pascanu}, \bibinfo{title}{{Embracing Change:
  Continual Learning in Deep Neural Networks}}, \bibinfo{year}{2020}.
  \DOIprefix\doi{10.1016/j.tics.2020.09.004}.
\bibitem[{Krenn et~al.(2022)Krenn, Pollice, Guo, Aldeghi, Cervera-Lierta,
  Friederich, dos Passos Gomes, H{\"{a}}se, Jinich, Nigam, Yao, and
  Aspuru-Guzik}]{Krenn2022OnScientific}
\bibinfo{author}{M.~Krenn}, \bibinfo{author}{R.~Pollice},
  \bibinfo{author}{S.~Y. Guo}, \bibinfo{author}{M.~Aldeghi},
  \bibinfo{author}{A.~Cervera-Lierta}, \bibinfo{author}{P.~Friederich},
  \bibinfo{author}{G.~dos Passos Gomes}, \bibinfo{author}{F.~H{\"{a}}se},
  \bibinfo{author}{A.~Jinich}, \bibinfo{author}{A.~K. Nigam},
  \bibinfo{author}{Z.~Yao}, \bibinfo{author}{A.~Aspuru-Guzik},
\newblock \bibinfo{title}{{On scientific understanding with artificial
  intelligence}},
\newblock \bibinfo{journal}{Nature Reviews Physics} \bibinfo{volume}{4}
  (\bibinfo{year}{2022}) \bibinfo{pages}{761--769}.
  \DOIprefix\doi{10.1038/s42254-022-00518-3}.
\bibitem[{Kitano(2021)}]{Kitano2021Nobel}
\bibinfo{author}{H.~Kitano}, \bibinfo{title}{{Nobel Turing Challenge: creating
  the engine for scientific discovery}}, \bibinfo{year}{2021}.
  \DOIprefix\doi{10.1038/s41540-021-00189-3}.
\bibitem[{Hitzler(2020)}]{Hitzler2020AReview}
\bibinfo{author}{P.~Hitzler},
\newblock \bibinfo{title}{{Semantic Web: A Review Of The Field}}
  (\bibinfo{year}{2020}). \URLprefix \url{https://doi.org/10.1145/nnnnnnn.}
  \DOIprefix\doi{10.1145/nnnnnnn}.
\bibitem[{Schoop et~al.(2006)Schoop, De~Moor, and {Dietz Jan L
  G}}]{Schoop2006Pragmatic}
\bibinfo{author}{M.~Schoop}, \bibinfo{author}{A.~De~Moor},
  \bibinfo{author}{{Dietz Jan L G}},
\newblock \bibinfo{title}{{pragmatic-web-2006-manifesto}},
\newblock \bibinfo{journal}{Communications of the ACM} \bibinfo{volume}{49}
  (\bibinfo{year}{2006}).
\bibitem[{Singh(2002)}]{Singh2002ThePragmatic}
\bibinfo{author}{M.~Singh},
\newblock \bibinfo{title}{{The Pragmatic Web}},
\newblock \bibinfo{journal}{IEEE Internet Computing} \bibinfo{volume}{6}
  (\bibinfo{year}{2002}) \bibinfo{pages}{4--5}. \URLprefix
  \url{http://computer.org/internet/}.
\bibitem[{Neiva et~al.(2016)Neiva, David, Braga, and Campos}]{Neiva2016Towards}
\bibinfo{author}{F.~W. Neiva}, \bibinfo{author}{J.~M.~N. David},
  \bibinfo{author}{R.~Braga}, \bibinfo{author}{F.~Campos},
\newblock \bibinfo{title}{{Towards pragmatic interoperability to support
  collaboration: A systematic review and mapping of the literature}},
\newblock \bibinfo{journal}{Information and Software Technology}
  \bibinfo{volume}{72} (\bibinfo{year}{2016}) \bibinfo{pages}{137--150}.
  \DOIprefix\doi{10.1016/j.infsof.2015.12.013}.
\bibitem[{Casanovas et~al.(2017)Casanovas,
  Rodr{\textbackslash}'{\textbackslash}iguez-Doncel, and
  Gonz{\'{a}}lez-Conejero}]{Casanovas2017TheRole}
\bibinfo{author}{P.~Casanovas},
  \bibinfo{author}{V.~Rodr{\textbackslash}'{\textbackslash}iguez-Doncel},
  \bibinfo{author}{J.~Gonz{\'{a}}lez-Conejero},
\newblock \bibinfo{title}{{The role of pragmatics in the web of data}},
\newblock in: \bibinfo{booktitle}{Pragmatics and Law},
  \bibinfo{publisher}{Springer}, \bibinfo{year}{2017}, pp.
  \bibinfo{pages}{293--330}. \DOIprefix\doi{10.1007/978-3-319-44601-1{\_}12}.
\bibitem[{Korb and Nicholson(2010)}]{Korb2010Bayesian}
\bibinfo{author}{K.~B. Korb}, \bibinfo{author}{A.~E. Nicholson},
  \bibinfo{title}{{Bayesian artificial intelligence}}, \bibinfo{publisher}{CRC
  press}, \bibinfo{year}{2010}.
\bibitem[{Lacave and D{\'{i}}ez(2002)}]{Lacave2002AReview}
\bibinfo{author}{C.~Lacave}, \bibinfo{author}{F.~J. D{\'{i}}ez},
  \bibinfo{title}{{A review of explanation methods for Bayesian networks}},
  \bibinfo{year}{2002}. \DOIprefix\doi{10.1017/S026988890200019X}.
\bibitem[{Tesic and Hahn(2021)}]{Tesic2021Explanation}
\bibinfo{author}{M.~Tesic}, \bibinfo{author}{U.~Hahn},
\newblock \bibinfo{title}{{Explanation in AI systems}},
\newblock in: \bibinfo{booktitle}{Human-Like Machine Intelligence},
  \bibinfo{publisher}{Oxford University Press}, \bibinfo{year}{2021}, pp.
  \bibinfo{pages}{114--136}.
  \DOIprefix\doi{10.1093/oso/9780198862536.003.0006}.
\bibitem[{{Briggs R A}(2014)}]{Briggs2014Normative}
\bibinfo{author}{{Briggs R A}}, \bibinfo{title}{{Normative theories of rational
  choice: Expected utility. Available online: }}, \bibinfo{year}{2014}.
  \URLprefix
  \url{https://plato.stanford.edu/entries/rationality-normative-utility/}.
\bibitem[{Fareh(2019)}]{Fareh2019Modeling}
\bibinfo{author}{M.~Fareh},
\newblock \bibinfo{title}{{Modeling incomplete knowledge of semantic web using
  Bayesian networks}},
\newblock \bibinfo{journal}{Applied Artificial Intelligence}
  \bibinfo{volume}{33} (\bibinfo{year}{2019}) \bibinfo{pages}{1022--1034}.
  \DOIprefix\doi{10.1080/08839514.2019.1661578}.
\bibitem[{Abaalkhail et~al.(????)Abaalkhail, Guthier, Alharthi, and
  Saddik}]{Abaalkhail2018Survey}
\bibinfo{author}{R.~Abaalkhail}, \bibinfo{author}{B.~Guthier},
  \bibinfo{author}{R.~Alharthi}, \bibinfo{author}{A.~E. Saddik},
  \bibinfo{title}{{Survey on Ontologies for Affective States and Their
  Influences}}, \bibinfo{type}{Technical Report}, ???? \URLprefix
  \url{http://tomdrummond.com/leading-and-caring-}.
\bibitem[{Barthes(2020)}]{Barthes2020Cognitive}
\bibinfo{author}{J.~P.~A. Barthes},
\newblock \bibinfo{title}{{Cognitive Agents and Ethical Behavior in
  Collaborative Teams}},
\newblock in: \bibinfo{booktitle}{Conference Proceedings - IEEE International
  Conference on Systems, Man and Cybernetics}, volume
  \bibinfo{volume}{2020-October}, \bibinfo{publisher}{Institute of Electrical
  and Electronics Engineers Inc.}, \bibinfo{year}{2020}, pp.
  \bibinfo{pages}{3776--3781}. \DOIprefix\doi{10.1109/SMC42975.2020.9282936}.
\bibitem[{Chari et~al.(2020{\natexlab{a}})Chari, Gruen, Seneviratne, and
  McGuinness}]{Chari2020Directions}
\bibinfo{author}{S.~Chari}, \bibinfo{author}{D.~M. Gruen},
  \bibinfo{author}{O.~Seneviratne}, \bibinfo{author}{D.~L. McGuinness},
\newblock \bibinfo{title}{{Directions for Explainable Knowledge-Enabled
  Systems}}  (\bibinfo{year}{2020}{\natexlab{a}}). \URLprefix
  \url{http://arxiv.org/abs/2003.07523}.
\bibitem[{Chari et~al.(2020{\natexlab{b}})Chari, Gruen, Seneviratne, and
  McGuinness}]{Chari2020Foundations}
\bibinfo{author}{S.~Chari}, \bibinfo{author}{D.~M. Gruen},
  \bibinfo{author}{O.~Seneviratne}, \bibinfo{author}{D.~L. McGuinness},
\newblock \bibinfo{title}{{Foundations of Explainable Knowledge-Enabled
  Systems}}  (\bibinfo{year}{2020}{\natexlab{b}}). \URLprefix
  \url{http://arxiv.org/abs/2003.07520}.
\bibitem[{Silver et~al.(2013)Silver, Yang, and
  Li}]{Silver2013LifelongAlgorithms}
\bibinfo{author}{D.~L. Silver}, \bibinfo{author}{Q.~Yang},
  \bibinfo{author}{L.~Li}, \bibinfo{title}{{Lifelong Machine Learning Systems:
  Beyond Learning Algorithms}}, \bibinfo{type}{Technical Report},
  \bibinfo{year}{2013}. \URLprefix \url{www.aaai.org}.
\bibitem[{van~de Ven et~al.(2022)van~de Ven, Tuytelaars, and
  Tolias}]{vandeVen2022ThreeLearning}
\bibinfo{author}{G.~M. van~de Ven}, \bibinfo{author}{T.~Tuytelaars},
  \bibinfo{author}{A.~S. Tolias},
\newblock \bibinfo{title}{{Three types of incremental learning}},
\newblock \bibinfo{journal}{Nature Machine Intelligence} \bibinfo{volume}{4}
  (\bibinfo{year}{2022}) \bibinfo{pages}{1185--1197}.
  \DOIprefix\doi{10.1038/s42256-022-00568-3}.
\bibitem[{Cerqueira et~al.(2020)Cerqueira, Torgo, and
  Mozeti{\v{c}}}]{Cerqueira2020EvaluatingMethods}
\bibinfo{author}{V.~Cerqueira}, \bibinfo{author}{L.~Torgo},
  \bibinfo{author}{I.~Mozeti{\v{c}}},
\newblock \bibinfo{title}{{Evaluating time series forecasting models: an
  empirical study on performance estimation methods}},
\newblock \bibinfo{journal}{Machine Learning} \bibinfo{volume}{109}
  (\bibinfo{year}{2020}) \bibinfo{pages}{1997--2028}.
  \DOIprefix\doi{10.1007/s10994-020-05910-7}.
\bibitem[{Lazaridou et~al.(2021)Lazaridou, Kuncoro, Gribovskaya, Agrawal,
  Li{\v{s}}ka, Terzi, Gimenez, de~Masson, Kocisky, Ruder, Yogatama, Cao, Young,
  and Blunsom}]{Lazaridou2021Mind}
\bibinfo{author}{A.~Lazaridou}, \bibinfo{author}{A.~Kuncoro},
  \bibinfo{author}{E.~Gribovskaya}, \bibinfo{author}{D.~Agrawal},
  \bibinfo{author}{A.~Li{\v{s}}ka}, \bibinfo{author}{T.~Terzi},
  \bibinfo{author}{M.~Gimenez}, \bibinfo{author}{C.~de~Masson},
  \bibinfo{author}{T.~Kocisky}, \bibinfo{author}{S.~Ruder},
  \bibinfo{author}{D.~Yogatama}, \bibinfo{author}{K.~Cao},
  \bibinfo{author}{S.~Young}, \bibinfo{author}{P.~Blunsom},
\newblock \bibinfo{title}{{Mind the Gap: Assessing Temporal Generalization in
  Neural Language Models}},
\newblock in: \bibinfo{booktitle}{Advances in Neural Information Processing
  Systems}, volume~\bibinfo{volume}{34}, \bibinfo{publisher}{Advances in Neural
  Information Processing Systems, 34, pp.29348-29363.}, \bibinfo{year}{2021},
  pp. \bibinfo{pages}{29348--29363}. \URLprefix \url{http://data.statmt.org/}.
\bibitem[{Bui et~al.(2022)Bui, Cho, and Yi}]{Bui2022Spatial}
\bibinfo{author}{K.~H.~N. Bui}, \bibinfo{author}{J.~Cho},
  \bibinfo{author}{H.~Yi},
\newblock \bibinfo{title}{{Spatial-temporal graph neural network for traffic
  forecasting: An overview and open research issues}},
\newblock \bibinfo{journal}{Applied Intelligence} \bibinfo{volume}{52}
  (\bibinfo{year}{2022}) \bibinfo{pages}{2763--2774}.
  \DOIprefix\doi{10.1007/s10489-021-02587-w}.
\bibitem[{Davidson and Moodley(2022)}]{Davidson2022ST-GNNs}
\bibinfo{author}{M.~Davidson}, \bibinfo{author}{D.~Moodley},
\newblock \bibinfo{title}{{ST-GNNs for Weather Prediction in South Africa}},
\newblock in: \bibinfo{booktitle}{Communications in Computer and Information
  Science}, volume \bibinfo{volume}{1734 CCIS}, \bibinfo{publisher}{Springer
  Science and Business Media Deutschland GmbH}, \bibinfo{year}{2022}, pp.
  \bibinfo{pages}{93--107}. \DOIprefix\doi{10.1007/978-3-031-22321-1{\_}7}.
\bibitem[{Dellermann et~al.(2019)Dellermann, Ebel, S{\"{o}}llner, and
  Leimeister}]{Dellermann2019HybridIntelligence}
\bibinfo{author}{D.~Dellermann}, \bibinfo{author}{P.~Ebel},
  \bibinfo{author}{M.~S{\"{o}}llner}, \bibinfo{author}{J.~M. Leimeister},
\newblock \bibinfo{title}{{Hybrid Intelligence}},
\newblock \bibinfo{journal}{Business and Information Systems Engineering}
  \bibinfo{volume}{61} (\bibinfo{year}{2019}) \bibinfo{pages}{637--643}.
  \DOIprefix\doi{10.1007/s12599-019-00595-2}.
\bibitem[{Gil(????)}]{GilThoughtfulDiscovery}
\bibinfo{author}{Y.~Gil}, \bibinfo{title}{{Thoughtful Artificial Intelligence:
  Forging A New Partnership for Data Science and Scientific Discovery}},
  \bibinfo{type}{Technical Report}, ????
\bibitem[{King et~al.(2009)King, Rowland, Aubrey, Liakata, Markham, Soldatova,
  Whelan, Clare, Young, Sparkes, and {others}}]{King2009TheAdam}
\bibinfo{author}{R.~D. King}, \bibinfo{author}{J.~Rowland},
  \bibinfo{author}{W.~Aubrey}, \bibinfo{author}{M.~Liakata},
  \bibinfo{author}{M.~Markham}, \bibinfo{author}{L.~N. Soldatova},
  \bibinfo{author}{K.~E. Whelan}, \bibinfo{author}{A.~Clare},
  \bibinfo{author}{M.~Young}, \bibinfo{author}{A.~Sparkes},
  \bibinfo{author}{{others}},
\newblock \bibinfo{title}{{The robot scientist Adam}},
\newblock \bibinfo{journal}{Computer} \bibinfo{volume}{42}
  (\bibinfo{year}{2009}) \bibinfo{pages}{46--54}.
\bibitem[{Gil et~al.(2016)Gil, Garijo, Ratnakar, Mayani, Adusumilli, Boyce, and
  Mallick}]{Gil2016AutomatedRepositories}
\bibinfo{author}{Y.~Gil}, \bibinfo{author}{D.~Garijo},
  \bibinfo{author}{V.~Ratnakar}, \bibinfo{author}{R.~Mayani},
  \bibinfo{author}{R.~Adusumilli}, \bibinfo{author}{H.~Boyce},
  \bibinfo{author}{P.~Mallick},
\newblock \bibinfo{title}{{Automated hypothesis testing with large scientific
  data repositories}},
\newblock in: \bibinfo{booktitle}{Proceedings of the Fourth Annual Conference
  on Advances in Cognitive Systems (ACS)}, volume~\bibinfo{volume}{2},
  \bibinfo{year}{2016}, p.~\bibinfo{pages}{4}.
\bibitem[{Soldatova et~al.(2013)Soldatova, Rzhetsky, De~Grave, and
  King}]{Soldatova2013RepresentationKnowledge}
\bibinfo{author}{L.~N. Soldatova}, \bibinfo{author}{A.~Rzhetsky},
  \bibinfo{author}{K.~De~Grave}, \bibinfo{author}{R.~D. King},
\newblock \bibinfo{title}{{Representation of probabilistic scientific
  knowledge}},
\newblock \bibinfo{journal}{Journal of Biomedical Semantics}
  \bibinfo{volume}{4} (\bibinfo{year}{2013}).
  \DOIprefix\doi{10.1186/2041-1480-4-S1-S7}.
\bibitem[{Ogundele et~al.(2016)Ogundele, Moodley, Pillay, and
  Seebregts}]{Ogundele2016AnOntology}
\bibinfo{author}{O.~Ogundele}, \bibinfo{author}{D.~Moodley},
  \bibinfo{author}{A.~Pillay}, \bibinfo{author}{C.~Seebregts},
\newblock \bibinfo{title}{{An ontology for factors affecting tuberculosis
  treatment adherence behavior in sub-Saharan Africa}},
\newblock \bibinfo{journal}{Patient Preference and Adherence}
  \bibinfo{volume}{10} (\bibinfo{year}{2016}).
  \DOIprefix\doi{10.2147/PPA.S96241}.
\bibitem[{Coetzer et~al.(2017)Coetzer, Moodley, and
  Gerber}]{Coetzer2017AKnowledge}
\bibinfo{author}{W.~Coetzer}, \bibinfo{author}{D.~Moodley},
  \bibinfo{author}{A.~Gerber},
\newblock \bibinfo{title}{{A knowledge-based system for generating interaction
  networks from ecological data}},
\newblock \bibinfo{journal}{Data {\&} Knowledge Engineering}
  \bibinfo{volume}{112} (\bibinfo{year}{2017}) \bibinfo{pages}{55--78}.
  \URLprefix
  \url{https://www.sciencedirect.com/science/article/pii/S0169023X17300459}.
  \DOIprefix\doi{10.1016/j.datak.2017.09.005}.
\bibitem[{Drake and Moodley(2022)}]{Drake2022INVEST:JSE}
\bibinfo{author}{R.~Drake}, \bibinfo{author}{D.~Moodley},
\newblock \bibinfo{title}{{INVEST: Ontology driven Bayesian networks for
  investment decision making on the JSE}},
\newblock in: \bibinfo{booktitle}{Proceedings of the Second Southern African
  Conference for Artificial Intelligence Research, 6-10 Dec. 2021, South
  Africa}, \bibinfo{year}{2022}, pp. \bibinfo{pages}{252--273}.
\bibitem[{Haig(2005)}]{Haig2005AnMethod}
\bibinfo{author}{B.~D. Haig},
\newblock \bibinfo{title}{{An abductive theory of scientific method}},
\newblock \bibinfo{journal}{Psychological Methods} \bibinfo{volume}{10}
  (\bibinfo{year}{2005}) \bibinfo{pages}{371--388}.
  \DOIprefix\doi{10.1037/1082-989X.10.4.371}.
\bibitem[{Haig(2018)}]{Haig2018AnMethod}
\bibinfo{author}{B.~D. Haig},
\newblock \bibinfo{title}{{An abductive theory of scientific method}},
\newblock in: \bibinfo{booktitle}{Method matters in psychology},
  \bibinfo{publisher}{Springer}, \bibinfo{year}{2018}, pp.
  \bibinfo{pages}{35--64}.
\bibitem[{Wanyana and Moodley(2021)}]{Wanyana2021AnEvolution}
\bibinfo{author}{T.~Wanyana}, \bibinfo{author}{D.~Moodley},
\newblock \bibinfo{title}{{An agent architecture for knowledge discovery and
  evolution}},
\newblock in: \bibinfo{booktitle}{German Conference on Artificial Intelligence
  (K{\"{u}}nstliche Intelligenz)}, \bibinfo{year}{2021}, pp.
  \bibinfo{pages}{241--256}. \DOIprefix\doi{10.1007/978-3-030-87626-5{\_}18}.
\bibitem[{Dardashti et~al.(2019)Dardashti, Hartmann, Th{\'{e}}bault, and
  Winsberg}]{Dardashti2019HawkingAnalysis}
\bibinfo{author}{R.~Dardashti}, \bibinfo{author}{S.~Hartmann},
  \bibinfo{author}{K.~Th{\'{e}}bault}, \bibinfo{author}{E.~Winsberg},
\newblock \bibinfo{title}{{Hawking radiation and analogue experiments: A
  Bayesian analysis}},
\newblock \bibinfo{journal}{Studies in History and Philosophy of Science Part B
  - Studies in History and Philosophy of Modern Physics} \bibinfo{volume}{67}
  (\bibinfo{year}{2019}) \bibinfo{pages}{1--11}.
  \DOIprefix\doi{10.1016/j.shpsb.2019.04.004}.
\bibitem[{Feldbacher-Escamilla and
  Gebharter(2020)}]{Feldbacher-Escamilla2020ConfirmationJeffrey}
\bibinfo{author}{C.~J. Feldbacher-Escamilla}, \bibinfo{author}{A.~Gebharter},
\newblock \bibinfo{title}{{Confirmation Based on Analogical Inference: Bayes
  Meets Jeffrey}},
\newblock \bibinfo{journal}{Canadian Journal of Philosophy}
  \bibinfo{volume}{50} (\bibinfo{year}{2020}) \bibinfo{pages}{174--194}.
  \DOIprefix\doi{10.1017/can.2019.18}.
\bibitem[{Aarons et~al.(2022)Aarons, Moodley, and
  Nzomo}]{Aarons2022AGeneralizable}
\bibinfo{author}{S.~Aarons}, \bibinfo{author}{D.~Moodley},
  \bibinfo{author}{M.~Nzomo}, \bibinfo{title}{{A Generalizable Hybrid Deep
  Learning Algorithm for the Detection of Atrial Fibrillation from Diverse
  Electrocardiogram Data}}, \bibinfo{type}{Technical Report}, University of
  Cape Town, \bibinfo{address}{Cape Town}, \bibinfo{year}{2022}. \URLprefix
  \url{https://projects.cs.uct.ac.za/honsproj/cgi-bin/view/2022/aarons_fisher_rosenthal.zip/deliverables/finalpapershai.pdf}.
\bibitem[{Russell and Norvig(2009)}]{Russell2009ArtificialApproach}
\bibinfo{author}{S.~Russell}, \bibinfo{author}{P.~Norvig},
  \bibinfo{title}{{Artificial intelligence: A modern approach}},
  \bibinfo{publisher}{Prentice-Hall}, \bibinfo{address}{Englewood Cliffs, NJ},
  \bibinfo{year}{2009}.
\bibitem[{Nzomo and Moodley(2023)}]{Nzomo2023Semantic}
\bibinfo{author}{M.~Nzomo}, \bibinfo{author}{D.~Moodley},
\newblock \bibinfo{title}{{Semantic Technologies in Sensor-Based Personal
  Health Monitoring Systems: A Systematic Mapping Study}}
  (\bibinfo{year}{2023}). \URLprefix \url{http://arxiv.org/abs/2306.04335}.
\bibitem[{Wanyana et~al.(2022)Wanyana, Nzomo, Price, and
  Moodley}]{Wanyana2022Combining}
\bibinfo{author}{T.~Wanyana}, \bibinfo{author}{M.~Nzomo},
  \bibinfo{author}{C.~S. Price}, \bibinfo{author}{D.~Moodley},
\newblock \bibinfo{title}{{Combining Machine Learning and Bayesian Networks for
  ECG Interpretation and Explanation}},
\newblock in: \bibinfo{booktitle}{International Conference on Information and
  Communication Technologies for Ageing Well and e-Health, ICT4AWE -
  Proceedings}, \bibinfo{publisher}{Science and Technology Publications, Lda},
  \bibinfo{year}{2022}, pp. \bibinfo{pages}{81--92}.
  \DOIprefix\doi{10.5220/0011046100003188}.
\bibitem[{Mudaly et~al.(2013)Mudaly, Moodley, Pillay, and
  Seebregts}]{Mudaly2013ArchitecturalCountries}
\bibinfo{author}{T.~Mudaly}, \bibinfo{author}{D.~Moodley},
  \bibinfo{author}{A.~Pillay}, \bibinfo{author}{C.~Seebregts},
\newblock \bibinfo{title}{{Architectural frameworks for developing national
  health information systems in low and middle income countries}},
\newblock in: \bibinfo{booktitle}{Proceedings of the 1st International
  Conference on Enterprise Systems, ES 2013}, \bibinfo{year}{2013}.
  \DOIprefix\doi{10.1109/ES.2013.6690083}.

\end{thebibliography}

\end{document}